\documentclass[runningheads]{llncs}

\usepackage{eccv}

\usepackage{eccvabbrv}

\usepackage{graphicx}
\usepackage{booktabs}
\usepackage{multirow}
\usepackage{graphicx}
\usepackage{bbding}
\usepackage{authblk}
\usepackage{algorithmic}
\usepackage{algorithm}
\usepackage{amsmath}

\usepackage[accsupp]{axessibility}  % Improves PDF readability for those with disabilities.

\usepackage{hyperref}

\usepackage{orcidlink}

\begin{document}

\title{Urban Waterlogging Detection: A Challenging Benchmark and Large-Small Model Co-Adapter}

\titlerunning{UW-Detection: A Challenging Benchmark and LSM Co-Adapter}

\author{Suqi Song\inst{1}$^\dagger$\orcidlink{0009-0007-6132-9212} \and
Chenxu Zhang\inst{1}$^\dagger$\orcidlink{0000-0002-7079-7284} \and
Peng Zhang\inst{1}\orcidlink{0009-0008-1123-0115} \and
Pengkun Li\inst{2} \and
Fenglong Song\inst{3} \and
Lei Zhang\inst{1}$^*$\orcidlink{0000-0002-5305-8543}}

\renewcommand{\thefootnote}{\fnsymbol{footnote}}
\footnotetext{$\dagger$ Equal contribution.}
\footnotetext{$*$ Corresponding author.}

\authorrunning{S. Song et al.}

\institute{School of Microelectronics and Communication
Engineering, Chongqing University, Chongqing 400044, China. \and
Huawei Technologies Co., Ltd., China \and
Huawei Noah's Ark Lab, Beijing 100196, China\\
\email{leizhang@cqu.edu.cn}}

\maketitle

\begin{abstract}
Urban waterlogging poses a major risk to public safety and infrastructure. Conventional methods using water-level sensors need high-maintenance to hardly achieve full coverage. Recent advances employ surveillance camera imagery and deep learning for detection, yet these struggle amidst scarce data and adverse environmental conditions. In this paper, we establish a challenging Urban Waterlogging Benchmark (UW-Bench) under diverse adverse conditions to advance real-world applications. We propose a Large-Small Model co-adapter paradigm (LSM-adapter), which harnesses the substantial generic segmentation potential of large model and the specific task-directed guidance of small model. Specifically, a Triple-S Prompt Adapter module alongside a Dynamic Prompt Combiner are proposed to generate then merge multiple prompts for mask decoder adaptation. Meanwhile, a Histogram Equalization Adap-ter module is designed to infuse the image specific information for image encoder adaptation. Results and analysis show the challenge and superiority of our developed benchmark and algorithm. \textit{Project page:} \url{https://github.com/zhang-chenxu/LSM-Adapter}
%constructed dataset demonstrate the effectiveness and superiority of the proposed LSM-adapter.
\end{abstract}

% ---------------------------------------------------------------
% Introduction
\section{Introduction}
\label{sec:intro}

%Urban waterlogging detection aims to identify areas with standing water in urban environments, including city streets, tunnels, and overpasses, through analysis of real data from water level sensors, images and video, etc.
Road water accumulation is a hidden danger, which not only causes structural damage to the pavement such as cracks and depressions, but also obstruct traffic flow, posing risks of accidents and threats to public safety. Therefore, early identification of waterlogged areas on urban roads is critical and essential.

Traditional urban waterlogging detection methods involve the installation of sensors on roadways to measure water levels, but challenges to maintain and hardly achieve full coverage \cite{sensor1, sensor2}. Recently, deep learning approaches have been explored for flood detection \cite{floodcnn1, floodcnn2, flood-maskrcnn, flooddata}, leveraging surveillance cameras. However, due to the lighting variability of water image reflection and the complexity of urban backgrounds, urban waterlogging detection faces several challenges: 1) Waterlogged areas vary in shape, size and depth, making it difficult to learn a uniform set of features; 2) The reflection on water surface along with shallow and clear standing water renders water texture information indistinct; 3) Under low-light conditions, the waterlogging features are not prominent, further intensifying the difficulty of detection. Owing to these challenges, existing methods struggle to detect waterlogging or provide accurate segmentation in real-world urban scenarios. Particularly, very limited scale and insufficient diversity of labeled data also diminish the generalizability of current methods, making urban waterlogging detection a hard nut to crack.

\begin{figure}[t]
  \centering
  \includegraphics[width=\columnwidth]{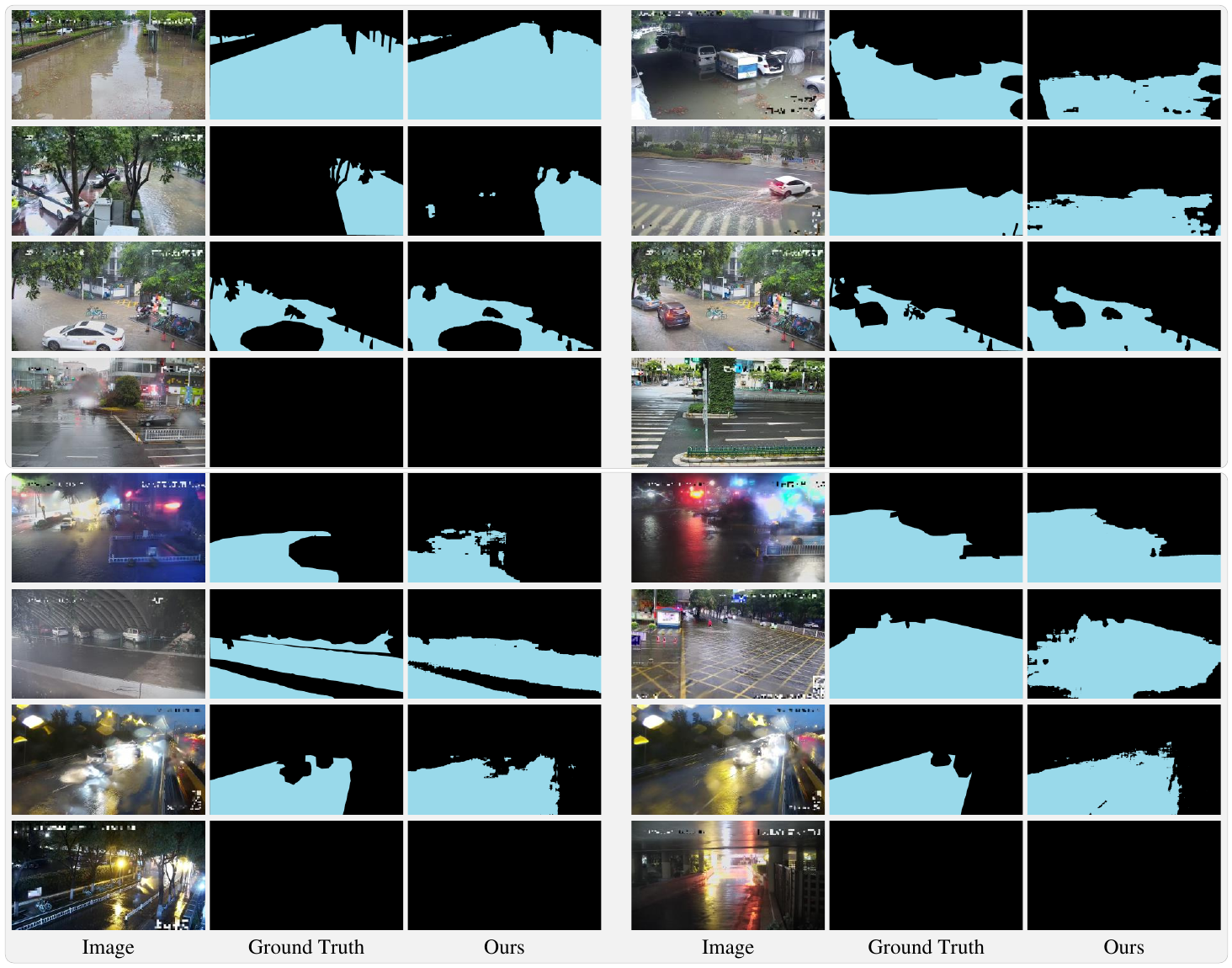}
  \caption{Waterlogging detection under general and hard conditions, such as strong-light reflection, low-light condition and clear water. The first 4 rows show general samples and hard samples for the last 4 rows. The practical difficulty of this task is witnessed.
  }
  \label{visual1}
\end{figure}

\begin{figure}[t]
  \centering
  \includegraphics[width=\columnwidth]{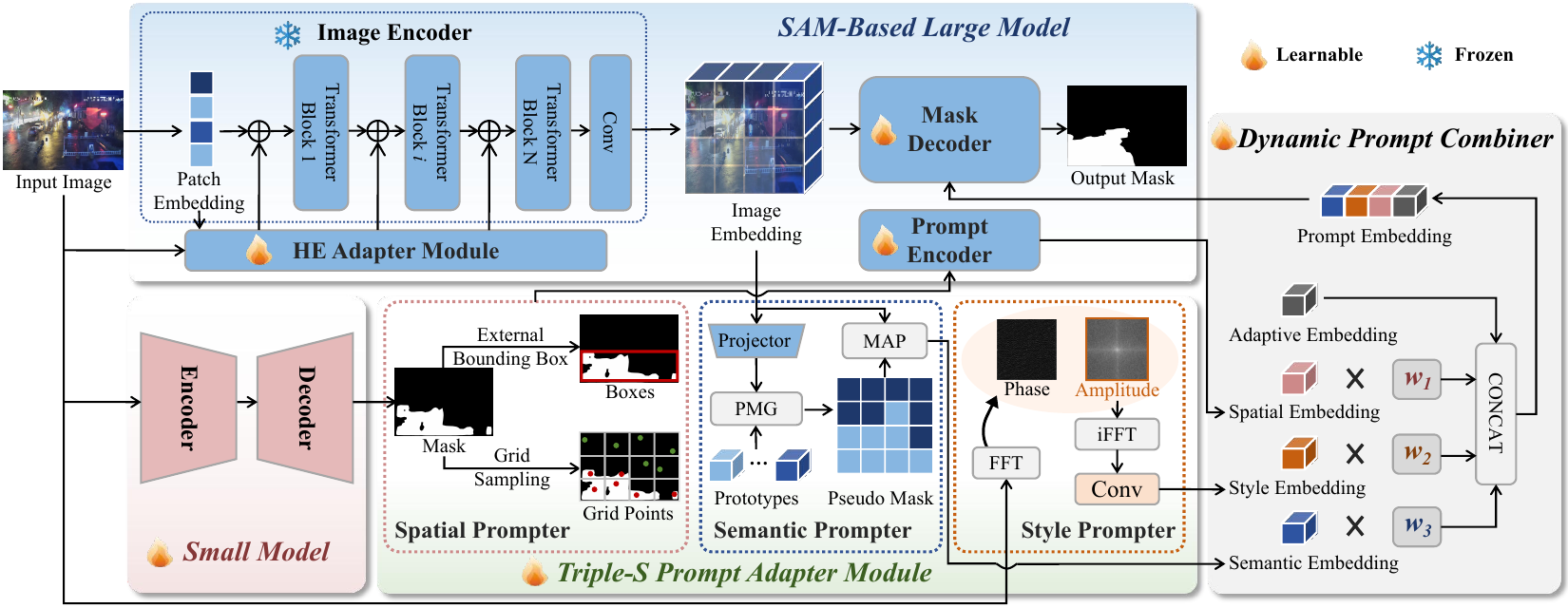}
  \caption{The proposed Large-Small Model Co-adapter Paradigm, which include a histogram equalization adapter, a triple-S prompt adapter and a dynamic prompt combiner. All components except the image encoder of SAM are trained for prompt generation, learning and adaptation, toward adverse waterlogging detection.
  }
  \label{framework}
\end{figure}

Recently, Meta AI has released an innovative visual foundation model known as the Segment Anything model (SAM) \cite{sam}. By prompt engineering and training on a corpus of over 1 billion images, SAM exhibits formidable zero-shot capabilities and impressive segmentation performance in numerous application fields \cite{samsurvey1}. However, due to lacking task-specific knowledge and reliance on manual prompts, SAM shows sub-optimal outcomes in downstream tasks \cite{samsurvey2}. Thereby, parameter fine-tuning \cite{medsam, polypsam}, integrating learnable adapters \cite{sam-adapter, sammed, cw-sam-adapter} or devising automated prompting \cite{autosam, rsprompter, clicksam, samac} appear to improve SAM in downstream tasks. But these techniques still have gaps with real-world waterlogging under adverse conditions. %Considering the robust generalization capabilities of SAM and its latent potential for various downstream tasks, it's worth investigating how to utilize SAM to aid urban waterlogging detection in addressing above challenges.

To advance urban waterlogging detection, the first challenge is data scarcity. %we propose a benchmark with a relatively large dataset with high-diversity and adequate evaluation protocols for urban waterlogging detection. The first
Existing datasets are of limited scale or lack diversity and comprise only samples that are easy to recognize \cite{flooddata}. Models trained on such data tend to exhibit poor generalizability, struggling to deploy in real-world application. To solve this practical issue, we firstly construct a challenging benchmark tailored for real-world urban waterlogging detection, including adverse conditions, such as low-light conditions, strong-light reflections and clear water, etc. A total of 7,677 waterlogging images are collected with manual labels, containing frames from surveillance cameras and handheld mobile devices. %Of these, there are 3,092 waterlogging images, while the remaining 4,585 images are not.
Fig.~\ref{visual1} shows the vision challenge of waterlogging images and the effectiveness of our approach.

To compromise the generalization of large model in diverse conditions and the specificity of small model in downstream task, we propose a SAM guided Large-Small Model co-adapter paradigm (LSM-adapter), exploring combined prompts tuning and adapting for efficient yet robust urban waterlogging detection. We design a Triple-S Prompt adapter (TSP-Adapt) comprising a small model based \underline{\textbf{s}}patial prompter, a prototype-based \underline{\textbf{s}}emantic prompter and a spectrum-based \underline{\textbf{s}}tyle prompter, to generate prompts from small model, large model and raw input, respectively. The origin and function of these prompts are distinct, to give complementary and counterbalancing benefits, thereby furnishing the large model with a more comprehensive and diverse information. Meanwhile, we propose a Dynamic Prompt Combiner (DPC) composed of a set of learnable weights and an adaptive embedding to dynamically weigh and blend the above prompts for mask decoder. Given the features of waterlogging images are often not prominent, we design a Histogram Equalization adapter (HE-Adpat) to infuse the enhanced task-relevant information (e.g., texture and contrast) into the image encoder. %The enhancement is achieved through adaptive histogram equalization, resulting in images with stronger contrast and clearer detail information.
The proposed LSM-adapter paradigm is overall elaborated in Fig. \ref{framework}.

In summary, our main contributions are as follows:
\begin{itemize}
    \item We first construct a challenging real-world urban waterlogging benchmark (UW-Bench) under adverse conditions, advancing the field towards application deployment with large model.
    \item We propose an innovative large-small model co-adapter paradigm (LSM-adapter), aiming at achieving win-win regime. In order to learn a robust prompter, a Triple-S prompt adapter (TSP-Adapt) with a dynamic prompt combiner is formulated, enabling a success on adaptation. %where the large model and the small model are interconnected and learn collaboratively through a triple-s multi-prompter module and a dynamic prompt combiner.
    \item We pioneer the use of vision foundation model i.e., SAM for urban waterlogging detection, providing new insights for future research.
\end{itemize}

% ---------------------------------------------------------------
% Related Work
\section{Related Work}
\subsection{Urban Waterlogging Detection}
Urban waterlogging detection is crucial for traffic management, urban planning, and disaster early warning systems. The early methods are based on water-level sensors\cite{sensor1, sensor2}, which detect water accumulation within a certain area through sensor devices placed at specific locations in a city. However, this approach is cost-ineffective to maintain and very limited in detection range. The remote sensing satellite imagery with a wide monitoring range are thus involved~\cite{remote1, remote2}. Due to the lack of local detail information in remote sensing-based methods, some studies have explored to utilize images or video data from surveillance cameras to detect waterlogging~\cite{video, flood, flood1, flood2, flood3}. \cite{video} combines local spatial-temporal features and brightness signals to detect water in videos by using decision forests. \cite{flood} estimates flood extent from crowdsourced images using brown color segmentation to identify flood water. More efforts were made to explore CNN based deep learning approaches \cite{floodcnn1, floodcnn2, flood-maskrcnn, flooddata}, such as Mask R-CNN \cite{maskrcnn} and DeepLabv3+\cite{deeplab}, and improved the waterlogging detection performance. In this paper, we pioneer the use of vision foundation model (SAM) with innovative designs on a newly developed urban waterlogging benchmark to advance this field fundamentally. %. focus on urban waterlogging semantic segmentation, combining the

\subsection{SAM Adaptation}
SAM~\cite{sam} is composed of a vision transformer-based image encoder, a lightweight mask decoder and a flexible prompt encoder that processes diverse input such as points, bounding boxes, masks and text. Numerous SAM variants have emerged, aiming to explore its potential in various tasks such as medical image analysis \cite{skinsam, polypsam, medsam, sammed}, camouflaged object detection \cite{samcod, sam-adapter} and mirror and transparent objects detection \cite{sammirror}. Adapting SAM to downstream tasks becomes a challenge. Early attempts involved directly fine-tuning a part of SAM (e.g., decoder) on downstream datasets \cite{skinsam, polypsam, medsam}. As full fine-tuning of image encoder is computationally intensive, some methods are inspired by adapters in natural language processing (NLP) and insert adapter in SAM, achieving efficient fine-tuning by training the adapter only~\cite{sam-adapter, cw-sam-adapter, sammed}. For example, SAM-Adapter~\cite{sam-adapter} adds adapters between transformer blocks of the image encoder. SAMed~\cite{sammed} employs the LoRA~\cite{lora} approach to approximate low-rank updates of the parameters in image encoder. Several studies accomplish adaptation from the perspective of generating automatic task-specific prompts~\cite{autosam, rsprompter, clicksam, samac}. For example, RSPrompter~\cite{rsprompter} generates appropriate prompts based on semantic information to yield semantically clear segmentation results for remote sensing images. In this paper, we consider dual adaptation in image- and prompt-level, and collaboration of large and small models.

% ---------------------------------------------------------------
% Method
\section{Method}

%\subsubsection{Framework Overview.}
%The proposed LSM-adapter paradigm is illustrated in \cref{framework}, comprising four essential components: 1) a SAM based task-generalized large model branch to provide pre-trained knowledge for segmentation, 2) a CNN based task-specific small model branch to automatically generate spatial prompt and injecting task-specific knowledge, 3) a Triple-S Prompter adapter (TSP-adapt) generating multi-level prompts beneficial to adaptation and segmentation, and 4) a Dynamic Prompt Combiner (DPC) to adaptively learn the final prompt embedding for mask decoder. %adjust the influence of each prompt and blend them.

\subsection{SAM-Based Task-Generalized Large Model Branch}
The SAM based large model is the main part in the entire framework for predicting the final segmentation mask. We retain three core components of SAM: the image encoder frozen with pretrained parameters, the lightweight mask decoder and the prompt encoder. As previously mentioned, directly deploying SAM to downstream task produces unsatisfactory results due to the frozen image encoder \cite{sam-adapter}. To facilitate the image encoder adaptation, we design an histogram equalization adapter laterally connected with the image encoder.
%Specifically, given an input image in this branch, it is fed into the SAM image encoder and the enhanced-image adapter module, with output as image embedding. A set of spatial prompts (points, boxes, or masks) serve as the input for prompt encoder to obtain spatial embedding, which is then combined with other embedding to derive prompt embedding (see \cref{sec:DPC} for details). The mask decoder is used to predict the segmentation mask with the dual inputs, the image embedding and prompt embedding.

\subsubsection{Histogram Equalization Adapter Module (HE-Adapt).}
\begin{figure}[tb]
  \centering
  \includegraphics[width=\columnwidth]{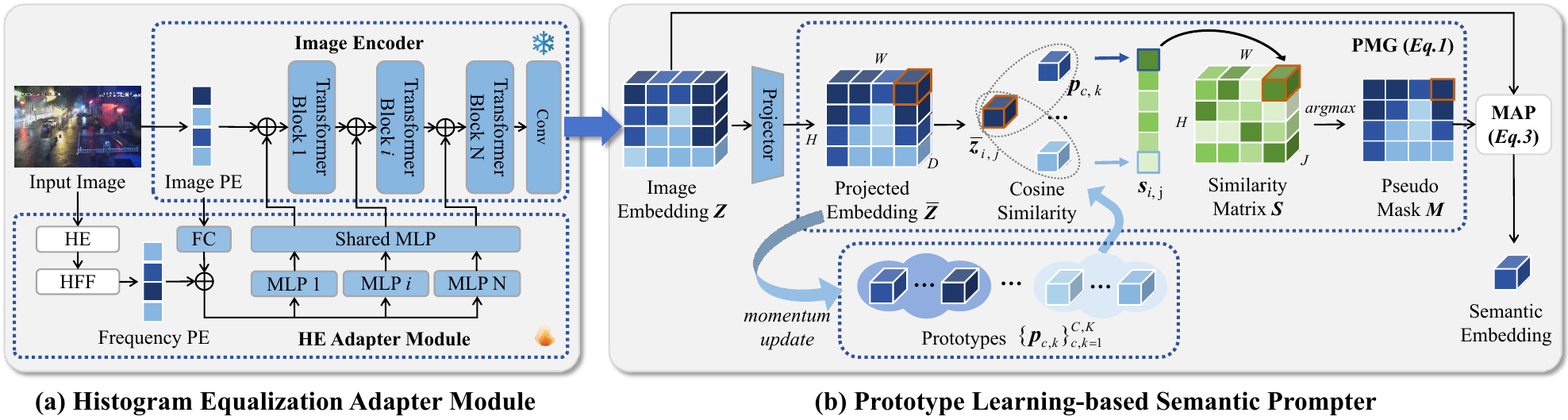}
  \caption{Details of the proposed histogram equalization adapter and the prototype learning based semantic prompter.
  }
  \label{HE-SemP}
\end{figure}

The internal structure of the enhanced-image adapter module is presented in \cref{HE-SemP} (a), which mainly consists of a histogram equalization, a high-frequency filter and MLP blocks. Given that the features of water are not pronounced in most challenging scenarios, we first conduct histogram equalization operation to highlight the contrast and texture of input image. %which can enhance the  of water, and make the boundaries more distinct.
The enhanced image is then passed through a high-frequency filter to extract high-frequency information beneficial for segmentation, and converted into frequency patch embedding. The patch embedding of original input image is reduced in dimension by fully-connected layer (FC) and added to the frequency patch embedding. This fused feature is mapped by $N$ individual MLP blocks and one parameter-shared MLP, and then merged with the original features of each transformer block in the SAM image encoder.

\subsection{CNN-Based Task-Specific Small Model Branch}
%The small model is a CNN-based semantic segmentation network.
Waterlogging possesses the reflectivity and transparency, making it easily camouflage itself due to lighting and complex environmental backgrounds. To this end, we adopt SiNet \cite{sinet}, that succeeds in camouflaged object detection, as our task-specific small model. To accommodate diverse requirements, the choice of the small model is flexible and can be substituted with any other network without necessitating alterations to the overarching framework.

Acting as a domain expert, small model interacts with the large model through spatial prompt, furnishing it with prior knowledge and directional task guidance. Given an input image, spatial prompt is generated by a spatial prompter built on the small model, which could be a predicted mask or a further processed version such as a bounding box or some points, encapsulating the spatial location information of the object to be detected (see \cref{sec:spa-prompter} for details).

\subsection{Triple-S Prompt Adapter Module}
The Triple-S prompt adapter module (TSP-Adapt) consists of a spatial prompter, a semantic prompter and a style prompter. %These components take the output from small model, the image embedding from large model and the raw input image, respectively, as their input. Then each prompter generates task-relevant and diverse prompts. These prompts, originating from different sources and serving different functions, complement and constrain each other, ensuring that the large model does not rely too heavily on any single prompt. %When one of the prompts contains insufficient or erroneous information, the others can play a corrective role.

\subsubsection{Spatial Prompter.}
\label{sec:spa-prompter}
SAM originally considered two sets of prompts, sparse (boxes or points) and dense (masks), both of which can provide spatial location information for the object to be segmented. We propose to generate such prompts via a spatial prompter utilizing the outputs of the small model. The masks $M_\mathrm{small}$ predicted by small model can be directly used as the dense prompts. Further processing of the masks $M_\mathrm{small}$ yields either boxes or points as sparse prompts. For box prompts, we take the bounding boxes of the regions composed of all pixels predicted as the foreground in the masks, represented by the coordinates of the top-left and bottom-right corners of the boxes. For point prompts, we divide the masks into multiple grid regions. In each grid area $G_{g\times{g}}$, all pixels are divided into a positive point set $I_P=\{(i,j)\mid{M_\mathrm{small}(i,j)}\geq{\tau}\}$ and a negative point set $I_N=\{(i,j)\mid{M_\mathrm{small}(i,j)}<{\tau}\}$, where $\tau$ is the preset threshold. If the set $I_P$ is not empty, we select one point $p\in{I_P}$ with the highest prediction confidence as the positive prompt and the prompt label is set to 1. Otherwise, we take one point $p\in{I_N}$ with the lowest prediction confidence as the negative prompt and the prompt label is set to 0. All points from the grids together form the grid point prompts, represented by their coordinates and labels. Despite providing three different types of prompts, considering the redundancy of information, we only select one prompt as the final spatial prompt fed into the prompt encoder and obtain spatial prompt $\boldsymbol{e}_{\mathrm{Spa}}$. It is noteworthy that the prompt encoder processes dense and sparse prompts differently. Dense prompts are embedded using convolution before adding to the image embedding and sparse prompts are encoded with positional encoding to generate the corresponding sparse embedding.

\subsubsection{Semantic Prompter.}
The image embedding of large model contains rich semantic information. Therefore, we propose a prototype learning-based semantic prompter, which leverages useful foreground features from large model to generate semantic prompts. The above process is detailed in \cref{HE-SemP} (b). A projector is first adopted to map the image embedding $\boldsymbol{Z}\in\mathbb{R}^{H\times{W}\times{D}}$ into the projected embedding $\boldsymbol{\bar{Z}}\in\mathbb{R}^{H\times{W}\times{D}}$. Inspired by \cite{prototype}, we randomly initialize a group of $C\cdot K$ prototypes, \textit{i.e.}, $\{\boldsymbol{p}_{c,k}\in\mathbb{R}^{D}\}^{C,K}_{c,k=1}$ in the embedding space, where $C$ is the number of categories and each class is represented by $K$ prototypes. For each pixel sample $\boldsymbol{\bar{z}}_{i,j}\in\mathbb{R}^D$, $i\in\{1\cdots W\}$, $j\in\{1\cdots H\}$ in the projected image embedding $\boldsymbol{\bar{Z}}$, we respectively calculate its cosine similarity with each prototype $\boldsymbol{p}_{c,k}$ to obtain a similarity vector $\boldsymbol{s}_{i,j}\in\mathbb{R}^{J}$ located at the ($i$, $j$) position in the similarity matrix $\boldsymbol{S}\in\mathbb{R}^{H\times{W}\times{J}}$, where $J=C\cdot K$. The category of the prototype corresponding to the maximum value   in the similarity vector $\boldsymbol{s}_{i,j}$ is assigned to the pixel sample $\boldsymbol{\bar{z}}_{i,j}$ as the pseudo label $c_{i,j}^*$. The pseudo mask generation (PMG) process can be represented as follows:
\begin{align}
  \boldsymbol{M}=\{c^{*}_{i, j}\}^{H,W}_{i,j=1},
  \text { with }\left(c^{*}_{i, j}, k^{*}_{i, j}\right)=\underset{(c, k)}{\arg \max }\left\{\langle\bar{z}_{i, j}, p_{c, k}\rangle\right\}_{c, k=1}^{C, K},
  \label{eq1}
\end{align}
where $\langle\cdot,\cdot\rangle$ denotes the cosine similarity operator. The pseudo mask $\boldsymbol{M}$ is then one-hot encoded and employed in conjunction with the original image embedding $\boldsymbol{Z}$ to compute the masked average pooling (MAP). This filters out irrelevant background features and preserves significant foreground features, and derives the semantic embedding as follows:%$\boldsymbol{e}_{\mathrm{Sem}}$:
\begin{align}
\boldsymbol{e}_{\mathrm{Sem}}=Concat(\boldsymbol{e}_{\mathrm{Sem}}^1, \boldsymbol{e}_{\mathrm{Sem}}^2,\cdots, \boldsymbol{e}_{\mathrm{Sem}}^C), %c\in\{1\cdots{C}\},
\label{eq2}
\end{align}
where $Concat(\cdot)$ denotes the concatenation operator and $\boldsymbol{e}_{\mathrm{Sem}}^c$ represents the semantic embedding of class $c$ ($c\in\{1\cdots{C}\}$), computed as follows:
\begin{align}
    \boldsymbol{e}_{\mathrm{Sem}}^c=\frac{\sum_{i,j}\boldsymbol{Z}(i,j)\odot\boldsymbol{M}^{c}(i,j)}{\sum_{i,j}\boldsymbol{M}^{c}(i,j)},
    \label{eq3}
\end{align}
where $\odot$ denotes the Hadamard product. The prototype $\boldsymbol{p}_{c,k}$ is momentum-updated after each training iteration according to the center of the $k$-th sub-cluster of the training samples assigned to the $c$-th class via online clustering. Meanwhile, a prototype loss $\mathcal{L}_{\mathrm{proto}}$ in \cite{prototype} is utilized to optimize the large model.

\subsubsection{Style Prompter.}
We introduce a spectrum-based style prompter that extracts image-specific style embedding from the input image as the third type of prompt. The style of an image refers to features such as color and texture. In the context of urban waterlogging detection, these features to some extent can reflect information about illumination and the scene, where illumination is a critical factor causing difficulty. Specifically, We first perform a 2D Fast Fourier Transform (FFT) on the input image $f(x,y)$ to acquire its frequency spectrum $F(u,v)$, which can be represented as:
\begin{align}
     F(u, v)=\text{FFT}\{f(x,y)\}=A(u, v) e^{j \Phi(u, v)}
     \label{eq4}
\end{align}
where $A(u,v)$ is the amplitude spectrum and $\Phi(u,v)$ is the phase spectrum. $u$ and $v$ are the frequency coordinates. Since the amplitude spectrum reflects the image style while the phase spectrum interprets the image content, we thus reconstruct the image solely by the amplitude spectrum using the 2D Inverse Fast Fourier Transform(iFFT):
\begin{align}
     \bar{f}(x,y)=\text{iFFT}\{A(u,v)\}
     \label{eq5}
\end{align}
The reconstructed amplitude-only image contains style information, which is then encoded as style embedding prompt $\boldsymbol{e}_{\mathrm{Sty}}$ by a convolutional block.

\subsection{Dynamic Prompt Combiner}
\label{sec:DPC}
The dynamic prompt combiner (DPC) is designed to find the optimal combination of the above three types of prompts. DPC comprises three sets of dynamic weights $\{w_1, w_2, w_3\}$ assigned to spatial, semantic and style prompt, respectively, and an adaptive embedding $\boldsymbol{e}_{Ada}$ learnable to improve potential bias. %First, the elements in the weight vector $\boldsymbol{W}=\{w_1, w_2, w_3\}$ are multiplicated with spatial, style, and semantic embedding $\{\boldsymbol{e}_{\mathrm{Spa}}, \boldsymbol{e}_{\mathrm{Sty}}, \boldsymbol{e}_{\mathrm{Sem}}\}$ correspondingly to assign them different weights.
%Thereafter, we introduce an adaptive embedding $\boldsymbol{e}_{Ada}$, which is a learnable embedding and is concatenated alongside all the aforementioned weighted embedding to form the final prompt embedding  $\boldsymbol{e}_{P}$.
The dynamically weighted prompts and the adaptive prompt are then concatenated to generate the final prompt as described in Fig. \ref{framework}, computed as follows:
\begin{align}
     \boldsymbol{e}_{\mathrm{P}}=Concat\{w_1\odot\boldsymbol{e}_{\mathrm{Spa}}, w_2\odot\boldsymbol{e}_{\mathrm{Sem}}, w_3\odot\boldsymbol{e}_{\mathrm{Sty}}, \boldsymbol{e}_{\mathrm{Ada}}\}.
     \label{eq6}
\end{align}
where $\odot$ denotes element-wise product.
During training, the weights are dynamically updated to encourage well-performing prompts while diminishing less-effective prompts. The motivation of the learnable embedding $\boldsymbol{e}_{Ada}$ arises from two aspects. 1) The learnable embedding enables the attention blocks within the decoder to comprehend nonlinear combination among these embeddings, and improve the bias that a linear combination may neglect. 2) It has a flexible capability to capture some useful implicit prompt information.

\begin{figure}[tb]
  \centering
  \includegraphics[width=\columnwidth]{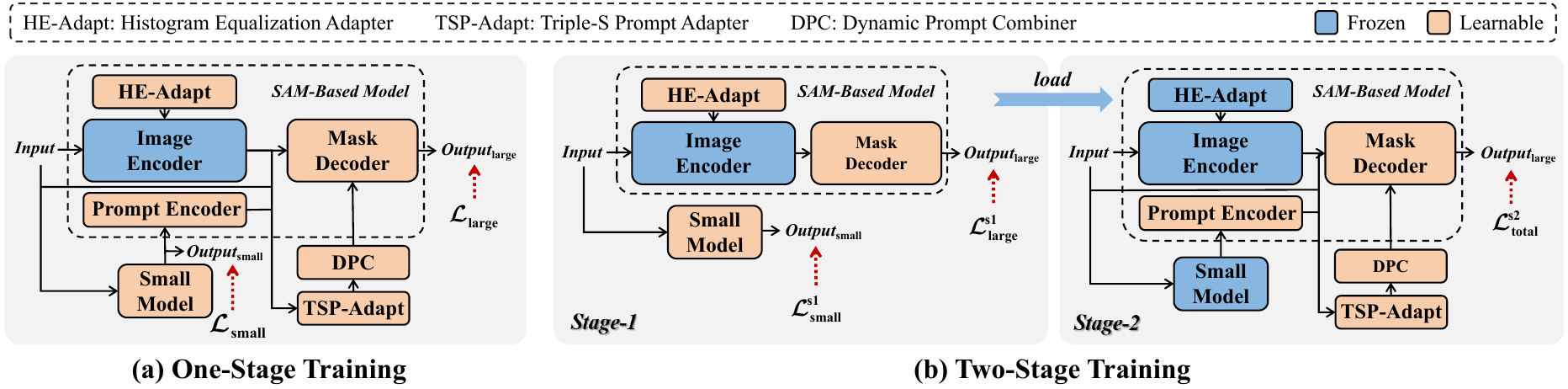}
  \caption{One-stage and Two-stage training strategies of the proposed large-small model paradigm for collaborative optimization.}
  \label{train}
\end{figure}

\subsection{Optimization}
\label{optimization_strategy}
Two training strategies are proposed to explore suitable joint training of models with diverse architectures, as illustrated in \cref{train}.

\subsubsection{One-stage Training.}
We introduce a straightforward one-stage training strategy, as depicted in \cref{train}~(a). The image encoder of the large model is frozen and the remaining parts are optimized together. We employ a combination of the focal loss $\mathcal{L}_{\mathrm{focal}}$~\cite{focal}, cross-entropy loss $\mathcal{L}_{\mathrm{ce}}$, IoU loss $\mathcal{L}_{\mathrm{iou}}$ and the prototype loss $\mathcal{L}_{\mathrm{proto}}$~\cite{prototype} for the large model:
\begin{align}
\mathcal{L}_{\mathrm{large}}=\mathcal{L}_{\mathrm{focal}}+\mathcal{L}_{\mathrm{ce}}+\mathcal{L}_{\mathrm{iou}}+\mathcal{L}_{\mathrm{proto}}.
\label{eq7}
\end{align}
The total loss is given as:
\begin{align}
    \mathcal{L}_{\mathrm{total}}=\mathcal{L}_{\mathrm{large}} + \lambda\mathcal{L}_{\mathrm{small}},
    \label{eq8}
\end{align}
where $\mathcal{L}_{\mathrm{small}}$ is the original loss of the small model (the loss function depends on specific model selection, and Mask-RCNN~\cite{maskrcnn}, U2Net~\cite{u2net} and SiNet \cite{sinet} are tested in experiments) and $\lambda$ is a hyper-parameter.% to adjust the influence of different training losses.

\subsubsection{Two-stage Training.}
A two-stage training strategy is provided to mitigate issues related to synchronization difficulties and gradient conflicts that may arise during the joint optimization of large-small models, as shown in \cref{train} (b).

In the first stage, the triple-s prompt adapter module, the dynamic prompt combiner and the prompt encoder are not involved. The image encoder remains frozen, while the remaining modules of the large model and small model are independently optimized by their own loss functions, i.e., $\mathcal{L}_{\mathrm{large}}^{\mathrm{s1}}$ and $\mathcal{L}_{\mathrm{small}}^{\mathrm{s1}}$, respectively. The loss function of small model is the same as Eq.~\ref{eq8}. The training loss of the large model for the first stage is defined as:
\begin{align}
\mathcal{L}_{\mathrm{large}}^{\mathrm{s1}}=\mathcal{L}_{\mathrm{focal}}+\mathcal{L}_{\mathrm{ce}}+\mathcal{L}_{\mathrm{iou}}.
\label{eq9}
\end{align}
In the second stage, we load the parameters of the modules (small model, HE-adapt and mask decoder) trained in the first stage, while integrating all the modules that were not considered previously for training (TSP-adapt, DPC and prompt encoder). With the parameters of image encoder, HE-adapt and the small model fixed, the optimization objective for the second stage is as follows:
\begin{align}
\mathcal{L}_{\mathrm{total}}^{\mathrm{s2}}=\mathcal{L}_{\mathrm{focal}}+\mathcal{L}_{\mathrm{ce}}+\mathcal{L}_{\mathrm{iou}}+\mathcal{L}_{\mathrm{proto}}.
\label{eq10}
\end{align}

% ---------------------------------------------------------------
% Experiments
\section{Experiments}
\subsection{Experimental Setup}
\subsubsection{Datasets.}
\begin{figure}[t]
  \centering
  \includegraphics[width=\columnwidth]{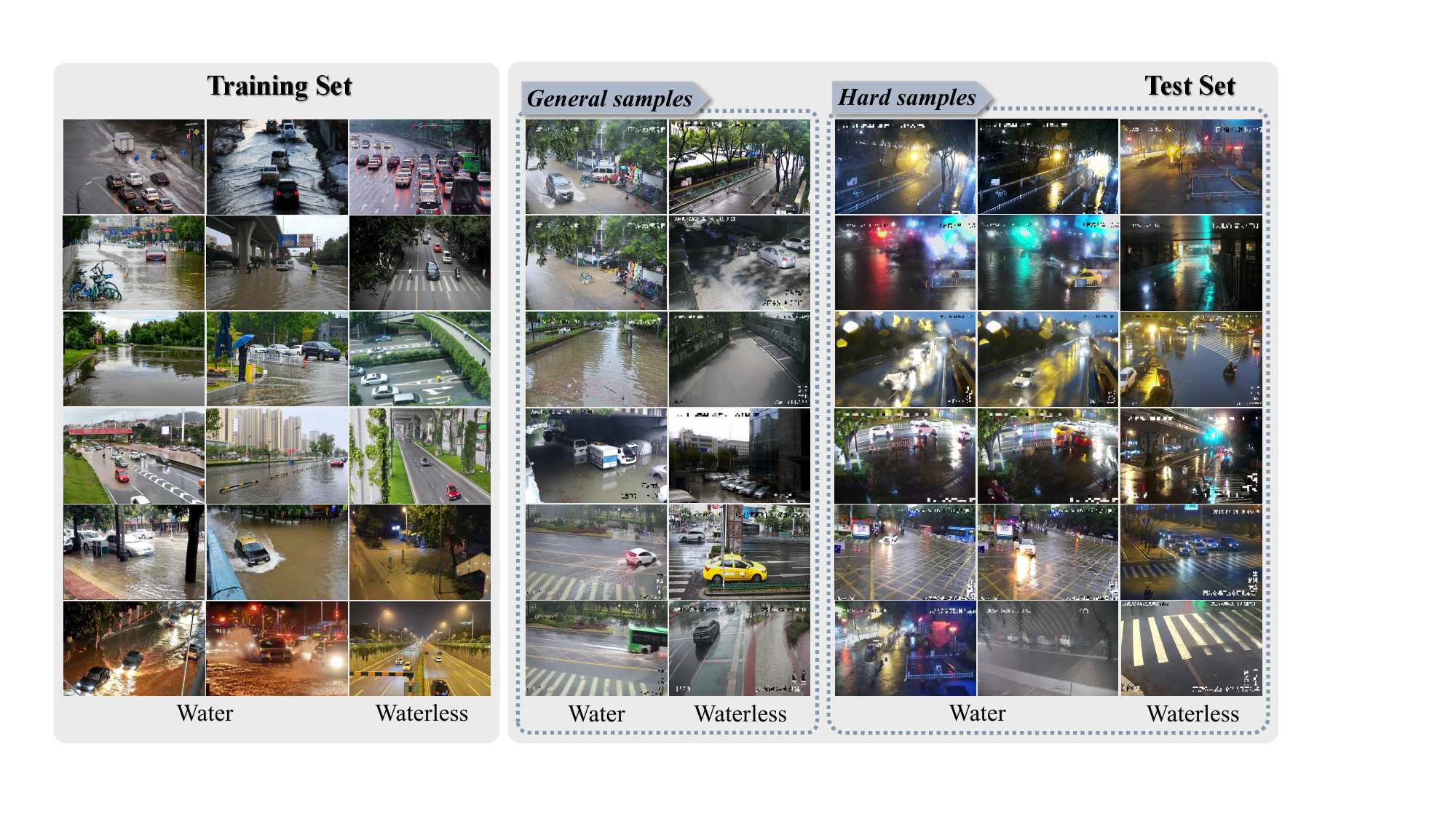}
  \caption{Training and testing examples in the developed UW-Bench. For objectively evaluating the capability of the model in real-world applications, we consider both \textit{general-sample} and \textit{hard-sample} cases in test set.}
  \label{dataset}
\end{figure}

For advancing urban waterlogging detection challenge, we develop a UW-Bench dataset containing a total of 7,677 images from various scenarios, such as waterlogging scenes, dry roads and hard cases, such as slippery roads and nighttime roads. Using keywords such as urban waterlogging, waterlogged roads, and monitoring viewpoints, we crawled and filtered relevant images from surveillance videos and handheld cameras. The training set includes 5,584 images, while the test set is provided by Huawei inc. and contains 2,093 images of urban scenes captured by surveillance cameras only. For the test set, we consider general-sample and hard-sample cases. Some examples from the training set and test set in our UW-Bench are described in Fig.~\ref{dataset}, which indicates the difficulty of detecting waterloggings.
In the labeling phase, we use \textit{EasyData} to annotate the dataset with masks. The pixel-level annotation process can be divided into several stages: training, annotation, validation, and correction. We first create some annotation samples and train the annotators to understand the annotation standard. We also assign an inspector to verify the mask annotations. For failed annotations, the inspector gives an explanation and feedback to each annotator to further improve the annotation quality. The overall annotation process ensures the accuracy and reliability of masks in waterlogging regions.

%Due to space limitation, we cannot provide more analysis about the dataset in this submission. The benchmark dataset and model with more details and protocols will be released publicly to advance this field.

%\begin{table}[t]
%\caption{Dataset statistics.}
%\label{tab:dataset}
%\centering
%\begin{tabular}{cc|cc|cc}
%\hline
%\multicolumn{2}{c|}{Training Set}
%& \multicolumn{4}{c}{Test Set} \\
%\hline
%\multirow{2}{*}{~ ~water~ ~} &\multirow{2}{*}{waterless~}
%& \multicolumn{2}{c|}{general} &\multicolumn{2}{c}{hard} \\
%\cline{3-6}
%& &~ ~water~ ~ &waterless~ &~ ~water~ ~ &waterless \\
%\hline
%\multicolumn{1}{c}{2,361} &\multicolumn{1}{c|}{3,223}
%&\multicolumn{1}{c}{} &\multicolumn{1}{c|}{} &\multicolumn{1}{c}{} & \multicolumn{1}{c}{}  \\
%\hline
%\end{tabular}
%\end{table}

\subsubsection{Evaluation Metrics.}
The waterlogging detection can be viewed as a pixel-level binary classification task for segmentation, and the waterlogged region is our prospect of interest. Based on the ground truth masks as well as the predicted waterlogging masks, we exploit the commonly used segmentation metrics, such as Precision, Recall, F1-score, and Intersection over Union (IoU) to evaluate the detection performance. %Precision-Recall curves (PR) are further presented for comparison. We also visualize the segmentation results on several general and hard samples.

\subsubsection{Implementation Details.}
For the large model, we choose the ViT-B version of the pre-trained SAM as the backbone, and the input image is resized to 1024*1024. In the training phase, the input images are randomly flipped horizontally, and the batch size is set as 2. AdamW optimizer is used with an initial learning rate of 0.0005, and cosine annealing decay is applied. In the testing phase, the final binary mask is obtained by a simple thresholding operation with the threshold set as 0.5. To evaluate our approach more comprehensively, we choose the classic Mask-RCNN~\cite{maskrcnn} for semantic segmentation, U2Net~\cite{u2net} for salient object detection and SINet~\cite{sinet} for camouflaged object detection, respectively, as the our task-specific small models, but not limited to these ones. All experiments are implemented in PyTorch on an NVIDIA Tesla V100S GPU (32G memory). See more implementation details in appendix.

\begin{table}[tb]
\caption{Comparison with existing methods on proposed UW-dataset. SAM-Adapter$^\dagger$ denotes training without prompts. The subscript of M, U, S denotes Mask R-CNN, U2Net and SINet, respectively, as the small model, to provide spatial prompts. The numbers in bold mean the best results.}
\label{tab:comparison}
\centering
\resizebox{\linewidth}{!}{
\begin{tabular}{l|cccc|cccc}
\hline
\multicolumn{1}{c|}{Test Set} &\multicolumn{4}{c|}{UW-all}         &\multicolumn{4}{c}{UW-hard} \\
\hline
\multicolumn{1}{c|}{Method} &Precision &~Recall &~F1-Score &~~IoU~~ & Precision &~Recall &~F1-Score &~~IoU~  \\
\hline
UNet~\cite{unet} &54.77 &45.58 &49.75 &33.12 &61.64 &30.40 &40.72 &25.56 \\
DeeplabV3~\cite{deeplabv3} &74.10 &47.17 &57.64 &40.50 &69.56 &36.04 &47.48 &31.13 \\
SETR\cite{SETR} &85.20 &54.01 &66.11 &49.37 &\textbf{81.71} &44.41 &57.54 &40.39 \\
Segformer\cite{segformer} &\textbf{86.63} &60.11 &70.98 &55.01 &81.22 &49.81 &61.75 &44.67 \\
Mask R-CNN~\cite{maskrcnn} &58.34 &51.86 &54.91 &37.84 &69.06 &42.15 &52.35 &35.46 \\
U2Net~\cite{u2net} &78.56 &49.86 &61.00 &43.89 &77.28 &39.89 &52.62 &35.70 \\
SINet\cite{sinet} &80.09 &59.02 &67.96 &51.47 &77.69 &52.00 &62.30 &45.24 \\
\hline	 	 			
SAM-Adapter$^\dagger$~\cite{sam-adapter} &72.13 &63.77 &67.69 &51.16 &69.70 &58.36 &63.53 &46.55 \\
SAM-Adapter$_{\mathrm{M}}$ &79.34 &60.94 &68.93 &52.60 &85.04 &58.49 &69.31 &53.03 \\
SAM-Adapter$_{\mathrm{U}}$ &80.63 &60.87 &69.37 &53.11 &77.63 &35.69 &48.90 &32.36 \\
SAM-Adapter$_{\mathrm{S}}$ &84.52 &61.25 &71.03 &55.07 &81.43 &54.57 &65.35 &48.53 \\
\hline	 	
LSM-Adapter$_\mathrm{M}$ &71.20 &\textbf{75.30} &73.19 &57.73 &73.39 &\textbf{74.16} &\textbf{73.77} &\textbf{58.45} \\
LSM-Adapter$_\mathrm{U}$ &74.99 &72.56 &73.75 &58.42 &75.02 &70.85 &72.88 &57.32 \\
LSM-Adapter$_\mathrm{S}$ &79.47 &70.57 &\textbf{74.76} &\textbf{59.69} &79.19 &67.29 &72.76 &57.18 \\
\hline
\end{tabular}}
\end{table}

\subsection{Experimental Results}
We evaluate the performance of the proposed LSM-Adapter on our developed UW-Bench under two types of test set: UW-all and UW-hard (a challenging subset of hard samples). We compare with some representative semantic segmentation models, including UNet~\cite{unet}, DeeplabV3~\cite{deeplabv3}, SETR~\cite{SETR}, Segformer~\cite{segformer}, Mask R-CNN~\cite{maskrcnn}, U2Net~\cite{u2net}, SINet~\cite{sinet} as well as SAM-Adapter~\cite{sam-adapter}, a large model based on SAM. %We employ the three small models mentioned above as the task-specific small models within our framework,
In the experiments, we adopt a two-stage training strategy in our LSM-Adapter and select the mask as the output type of the spatial prompter (notably, experiments on different training strategies and spatial prompt types are discussed in Section~\ref{discussion_analysis}). Additionally, SAM-Adapter utilizes a default prompt embedding as one of the dual inputs of the mask decoder and the prompt encoder was omitted. For a fair comparison, we also feed the prompts into SAM-Adapter based on the three small models, respectively, following the same setting.

The quantitative comparisons are tabulated in \cref{tab:comparison}. From the results, we witness our proposed method achieves state-of-the-art performance in both test sets, and significantly outperforms extant methodologies, particularly in Recall, F1 score and IoU under different small models.  Specifically, LSM-Adapter$_\mathrm{M}$, LSM-Adapter$_\mathrm{U}$, and LSM-Adapter$_\mathrm{S}$ demonstrate increment by 6.8$\%$ to 18.28$\%$ in F1 score and 8.22$\%$ to 19.89$\%$ in IoU, compared with their respective small models. Particularly, LSM-Adapter$_\mathrm{M}$ exhibits an increment of 19.89$\%$ over Mask R-CNN on IoU, indicating that small models with inferior standalone performance are capable of realizing more pronounced performance improvements if co-trained with large model. Compared to the large model i.e. SAM-Adapter, our approach is improved by 7.45$\%$ to 9.02$\%$ in F1 score and 6.57$\%$ to 8.53$\%$ in IoU. Moreover, the competitive small model, i.e., SINet, exhibits an even greater gain in overall performance when integrated with the large model.

\begin{figure}[tb]
  \centering
  \begin{subfigure}{0.32\linewidth}
    \includegraphics[width=\columnwidth]{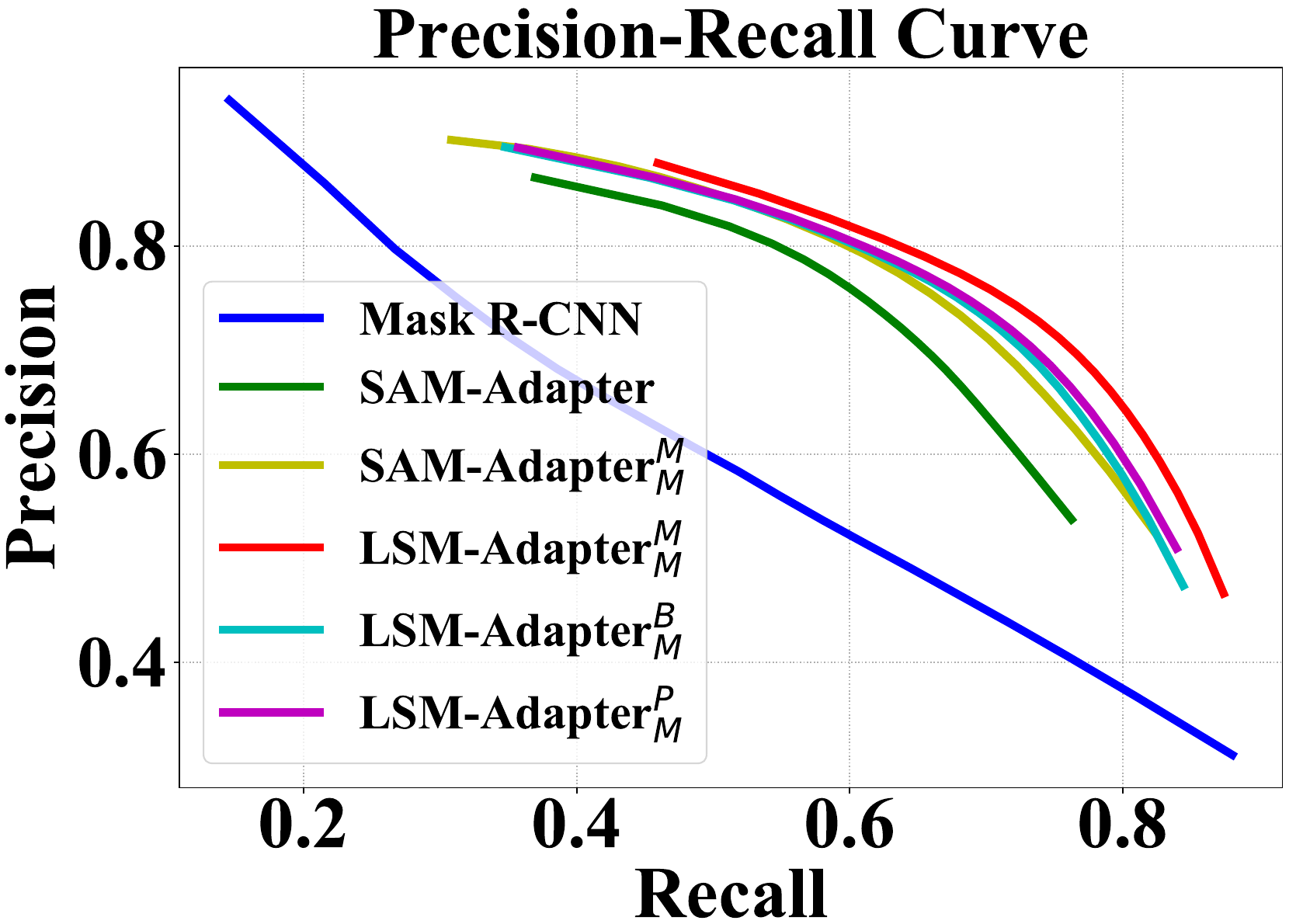}
    \caption{}
    \label{PR1}
  \end{subfigure}
  \hfill
  \begin{subfigure}{0.32\linewidth}
    \includegraphics[width=\columnwidth]{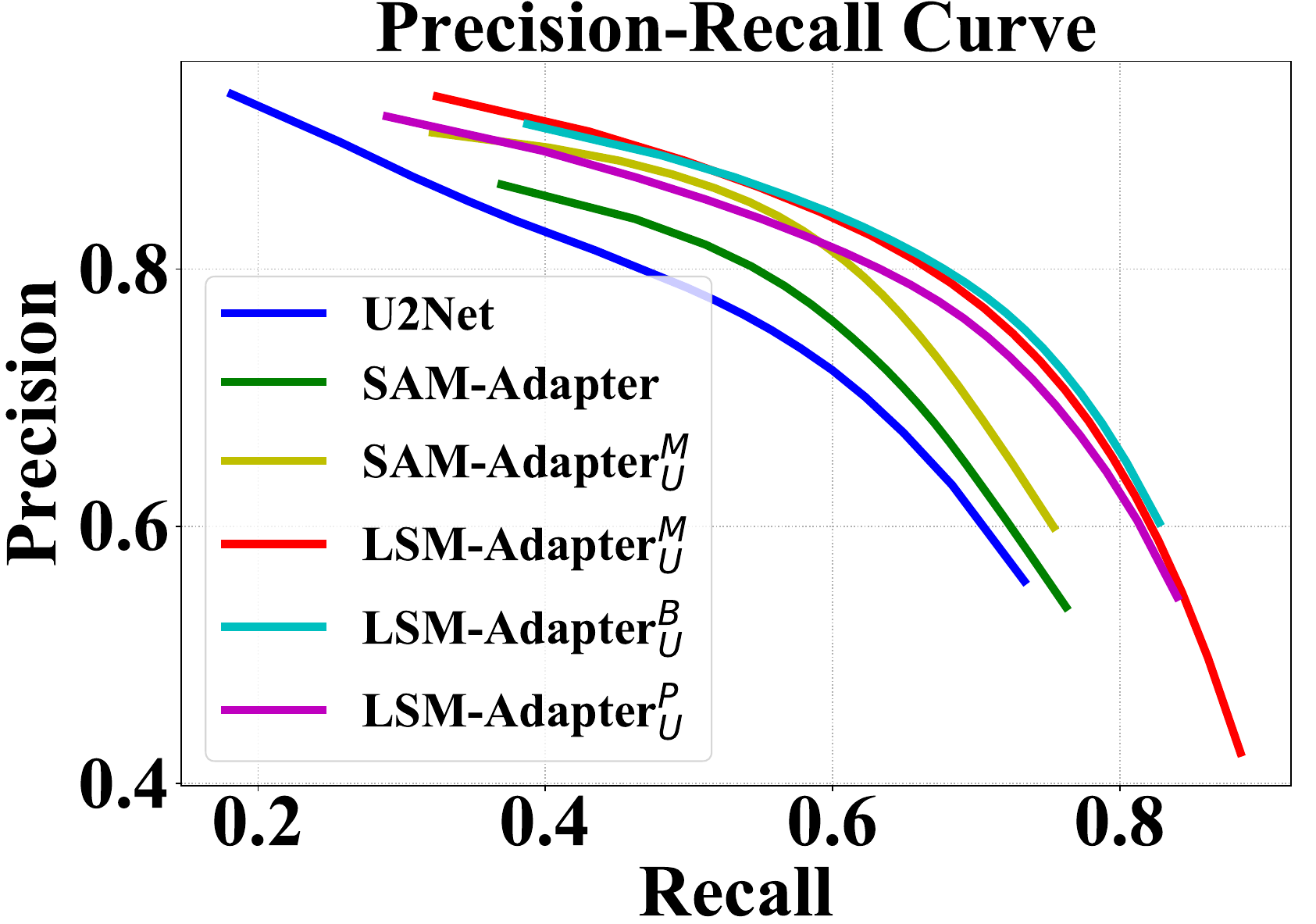}
    \caption{}
    \label{PR2}
  \end{subfigure}
  \hfill
  \begin{subfigure}{0.32\linewidth}
    \includegraphics[width=\columnwidth]{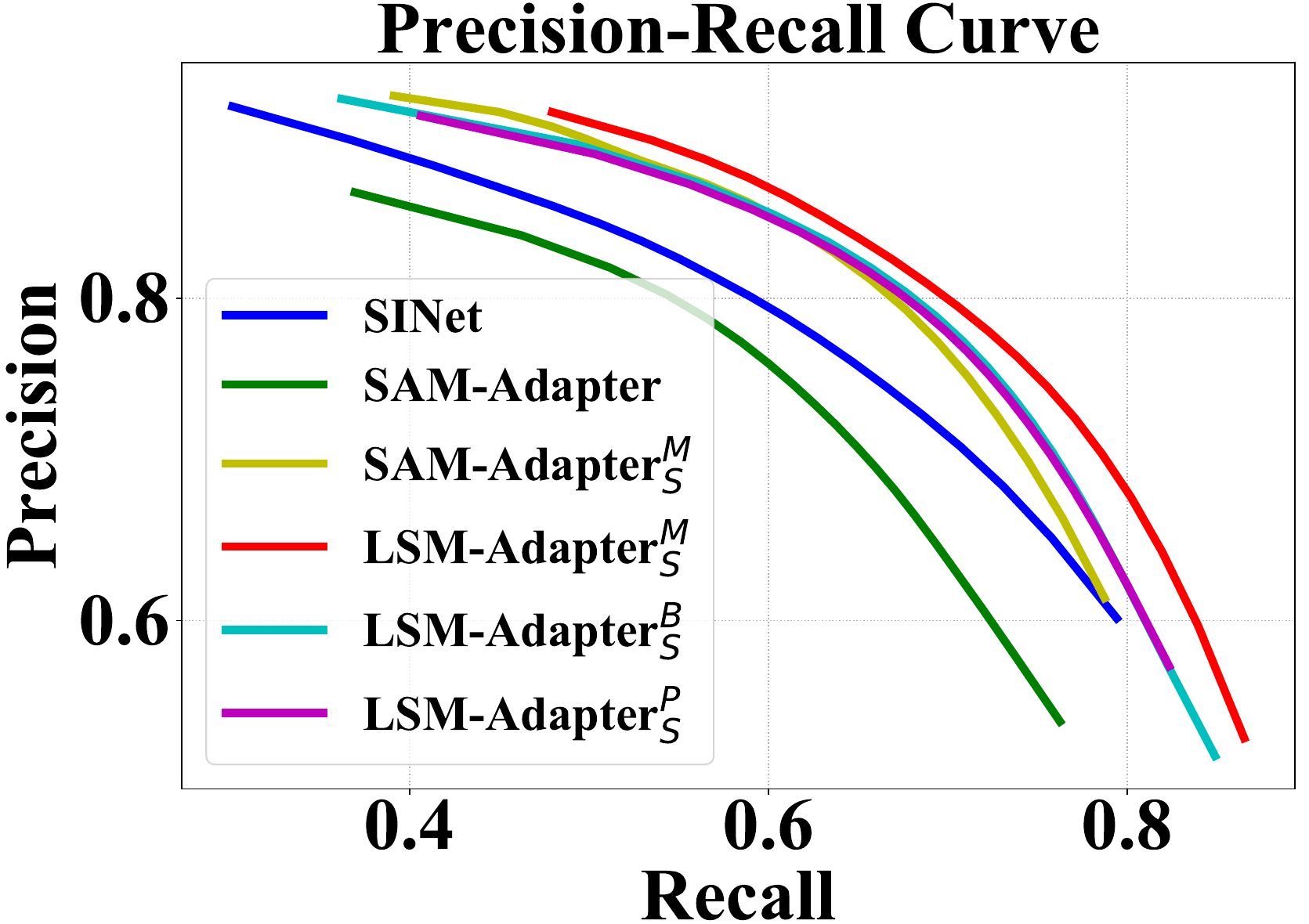}
    \caption{}
    \label{PR3}
  \end{subfigure}
  \caption{Precision-Recall curves of our models and other existing methods. M, B, P denotes mask, box and point is used as spatial prompt, respectively. M, U, S denotes Mask-RCNN (a), U2Net (b) and SINet (c) is used as small model, respectively.}
  \label{PR}
\end{figure}

For the purpose of qualitative analysis, we illustrate the waterlogging segmentation results on several general test samples and hard test samples in \cref{visual1}. Evidently, the predicted masks of LSM-Adapter better approach the ground truth, further demonstrating its superiority to other methods. We further exploit the precision-recall curves (PR) to compare different methods. \cref{PR} illustrates the PR curves of our methods and other existing methods. Each subplot corresponds to the use of different small models. In each subplot, the PR-cureve of our method is closer to the top-right corner and exhibits better performance than existing CNN based segmentation models and Transformer based SAM-Adapter.
%, demonstrating the superiority of our approach.

\begin{table}[]
  \caption{Effects of different training strategies and spatial prompts. 1-S and 2-S denotes the one-stage and two-stage training strategy, respectively, as discussed in Section~\ref{optimization_strategy}. The spatial prompts include mask, box and point. For each model, the numbers in bold mean the best results across the same training strategy.}
  \label{tab:analysis}
  \centering
  \scalebox{0.9}{
  \begin{tabular}{c|c|c|p{2.0cm}p{2.0cm}p{2.0cm}p{2.0cm}}
    \hline
    Method &\multicolumn{1}{l|}{~Train~~} &\multicolumn{1}{l|}{~Prompt~~} &\multicolumn{1}{c}{Precision} &\multicolumn{1}{c}{Recall} &\multicolumn{1}{c}{F1-Score} &\multicolumn{1}{c}{IoU}   \\ \hline
    % LSM-Adapt(Mask R-CNN)
    \multirow{6}{*}{LSM-Adapter$_\mathrm{M}$}
    &\multirow{3}{*}{1-S}      	 	
    &Mask &\multicolumn{1}{c}{73.06} &\multicolumn{1}{c}{53.71} &\multicolumn{1}{c}{\textbf{61.91}} &\multicolumn{1}{c}{\textbf{44.86}}  \\ &
    &Box &\multicolumn{1}{c}{77.80} &\multicolumn{1}{c}{51.01} &\multicolumn{1}{c}{61.62} &\multicolumn{1}{c}{44.53}  \\ &
    &Point &\multicolumn{1}{c}{68.11} &\multicolumn{1}{c}{47.04} &\multicolumn{1}{c}{55.65} &\multicolumn{1}{c}{38.55}  \\ \cline{2-7}
    &\multirow{3}{*}{2-S}
    &Mask &\multicolumn{1}{c}{71.20} &\multicolumn{1}{c}{75.30} &\multicolumn{1}{c}{\textbf{73.19}} &\multicolumn{1}{c}{\textbf{57.73}}  \\ &
    &Box &\multicolumn{1}{c}{73.60} &\multicolumn{1}{c}{69.46} &\multicolumn{1}{c}{71.47} &\multicolumn{1}{c}{55.61}  \\ &
    &Point &\multicolumn{1}{c}{74.73} &\multicolumn{1}{c}{68.85} &\multicolumn{1}{c}{71.67} &\multicolumn{1}{c}{55.85}  \\ \hline
    % LSM-Adapt(U2Net)
    \multirow{6}{*}{LSM-Adapter$_\mathrm{U}$}
    &\multirow{3}{*}{1-S}
    &Mask &\multicolumn{1}{c}{66.84} &\multicolumn{1}{c}{66.34} &\multicolumn{1}{c}{66.59} &\multicolumn{1}{c}{49.94}  \\ &
    &Box &\multicolumn{1}{c}{72.21} &\multicolumn{1}{c}{63.75} &\multicolumn{1}{c}{\textbf{69.01}} &\multicolumn{1}{c}{\textbf{52.68}}  \\ &      	 	
    &Point &\multicolumn{1}{c}{71.65} &\multicolumn{1}{c}{54.96} &\multicolumn{1}{c}{62.20} &\multicolumn{1}{c}{45.14}  \\ \cline{2-7}
    &\multirow{3}{*}{2-S}
    &Mask &\multicolumn{1}{c}{74.99} &\multicolumn{1}{c}{72.56} &\multicolumn{1}{c}{73.75} &\multicolumn{1}{c}{58.42}  \\ &
    &Box &\multicolumn{1}{c}{78.98} &\multicolumn{1}{c}{69.22} &\multicolumn{1}{c}{\textbf{73.78}} &\multicolumn{1}{c}{\textbf{58.45}}  \\ &
    &Point &\multicolumn{1}{c}{77.51} &\multicolumn{1}{c}{67.36} &\multicolumn{1}{c}{72.08} &\multicolumn{1}{c}{56.35}  \\ \hline
    % LSM-Adapt(SINet)
    \multirow{6}{*}{LSM-Adapter$_\mathrm{S}$}
    &\multirow{3}{*}{1-S}
    &Mask &\multicolumn{1}{c}{63.88} &\multicolumn{1}{c}{62.58} &\multicolumn{1}{c}{\textbf{63.22}} &\multicolumn{1}{c}{\textbf{46.22}}  \\ &
    &Box &\multicolumn{1}{c}{61.84} &\multicolumn{1}{c}{55.27} &\multicolumn{1}{c}{58.37} &\multicolumn{1}{c}{41.22}  \\ &
    &Point &\multicolumn{1}{c}{63.60} &\multicolumn{1}{c}{60.58} &\multicolumn{1}{c}{62.05} &\multicolumn{1}{c}{44.98}  \\ \cline{2-7}
    &\multirow{3}{*}{2-S}
    &Mask &\multicolumn{1}{c}{79.47} &\multicolumn{1}{c}{70.57} &\multicolumn{1}{c}{\textbf{74.76}} &\multicolumn{1}{c}{\textbf{59.69}}  \\ &
    &Box &\multicolumn{1}{c}{75.75} &\multicolumn{1}{c}{72.21} &\multicolumn{1}{c}{73.94} &\multicolumn{1}{c}{58.65}  \\ &
    &Point &\multicolumn{1}{c}{78.00} &\multicolumn{1}{c}{69.85} &\multicolumn{1}{c}{73.70} &\multicolumn{1}{c}{58.35}  \\ \hline
  \end{tabular}}
\end{table}

\subsection{Discussion on Training Strategies and Spatial Prompts}
\label{discussion_analysis}
We explore the impact of employing different training strategies and spatial prompts on model performance. The results are presented in \cref{tab:analysis}.
\subsubsection{Results based on different training strateges.}
The proposed LSM-Adapter that employs the one-stage training strategy is significantly inferior to the two-stage training strategy by comparing their best performances. This demontrates the proposed two-stage training strategy is more stable in adaptation.
%For comparison of different spatial prompts, improvements are only observed when the box is used as the spatial prompt. Specifically, LSM-Adapter$_\mathrm{U}$ demonstrates improved performance compared to both SAM-Adapter and U2Net.
We posit that the following factors may impede the effective implementation of the one-stage training strategy. During the early stages of training, the predicted output of the small model is characterized by low accuracy, and add complexity to the training of the large model results in subsequent slow convergence. Concurrently, the joint training of two networks with distinct architectures is highly contingent upon the selection of suitable hyper-parameters to achieve a synchronized optimization process. Otherwise, the optimization objectives may conflict each other, thereby hindering the joint model from attaining the optimal performance.

\subsubsection{Results based on different spatial prompts.}
Under identical conditions concerning the small model and training strategy, we compare three types of spatial prompts. Except LSM-Adapter$_\mathrm{U}$, the performance by using mask as spatial prompt consistently surpasses that of other two types of prompts (box and point) in all scenarios. Although the box prompt in LSM-Adapter$_\mathrm{U}$ is the best, the performance gap from the mask prompt is very small, with a mere 0.03$\%$ gap for both F1-score and IoU when employing two-stage training strategy. Moreover, the performance of models employing mask prompts predominantly exceeds that of both large models and their respective small models. A possible explanation for this observation may be that, in comparison to the sparse prompts of boxes and points, mask prompts furnish a more abundant referential information.

\begin{table}[t]
\caption{Ablation studies for innovative components. SAM-Adapter is underlined.}
\label{tab:ablation}
\centering
\scalebox{0.9}{
\begin{tabular}{ccccc|cc|cc}
\hline
\multirow{2}{*}{HE-Adapt} & \multirow{2}{*}{SpaP} & \multirow{2}{*}{SemP} & \multirow{2}{*}{StyP} & \multirow{2}{*}{DPC} & \multicolumn{2}{c|}{~~~LSM-Adapter$_\mathrm{U}$~~~} & \multicolumn{2}{c}{~~~LSM-Adapter$_\mathrm{S}$~~~}  \\
\cline{6-9}
& & & & &F1-Score &IoU &F1-Score &IoU \\
\hline
& & & &    	 	 			
&\underline{65.74} &\underline{51.16} &\underline{65.74} &\underline{51.16} \\
\CheckmarkBold & & & &     	
&71.65 &55.83 &71.65 &55.83 \\
\CheckmarkBold &\CheckmarkBold & & &
&71.97 &56.22 &73.95 &58.67 \\
\CheckmarkBold &\CheckmarkBold &\CheckmarkBold & &
&72.16 &56.45 &73.98 &58.71 \\
\CheckmarkBold &\CheckmarkBold &\CheckmarkBold &\CheckmarkBold &
&72.72 &57.14 &74.63 &59.52 \\
\CheckmarkBold &\CheckmarkBold &\CheckmarkBold &\CheckmarkBold &\CheckmarkBold
&\textbf{73.75} &\textbf{58.42} &\textbf{74.76} &\textbf{59.69} \\
\hline
\end{tabular}}
\end{table}

\subsection{Ablation Study}
To verify the effectiveness of each module in our proposed framework, we adopt the two-stage training strategy, select mask as the spatial prompt and exploit $\mathrm{LSM}_\mathrm{U2Net}$ and $\mathrm{LSM}_\mathrm{SINet}$ to conduct the ablation experiments. SAM-Adapter is the baseline model, i.e., without adding any of our proposed modules. Experimental results are shown in \cref{tab:ablation}. %Note that the numbers in underline denote the results of SAM-Adapter (baseline).
We can see that adding adaptive histogram equalization adapter (HE-adapt) achieves performance improvement in F1-score and IoU, which proves the effect of HE-adapt for image encoder adaptation. %operation to the task of waterlogging detection.
Additionally, with spatial prompt (SpaP), semantic prompt (SemP), and style prompt (StyP) are added gradually, the performance is also improved gradually, which proves the effectiveness of the Triple-S prompt adapter (TSP-adapt) for mask decoder adaptation. %prompts obtained from the small model, the large model encoder, and the input image, respectively.
On this basis, by exploiting the dynamic prompt combiner (DPC), the optimal performance is achieved, which proves our proposed DPC can effectively combine different prompts via a dynamic optimization strategy to improve potential bias implied in an individual prompt. %more reasonably and effectively than the simple combination.

\section{Conclusion and Outlook}
In this paper, we first pioneer the use of large vision model (i.e., SAM) for a challenging downstream task i.e. urban waterlogging detection and advancing its real-world applications. Considering the generic segmentation capability of large model and the task-specificity of small model, we propose a large-small model co-adapter paradigm following the win-win mechanism. To address the data scarcity of real-world waterlogging detection, we further contribute a large benchmark to advance this field fundamentally, on which more powerful algorithms can be developed.
In experiments, we provide new perspectives on the training strategy of large-small model collaboration, due to their architectural differences. %We also reveal what kind of spatial prompt is better suitable to the waterlogging segmentation task.
This paper sheds light on the possibility of large model in adapting to challenging downstream task with congenital data scarcity and adverse conditions.

In the future work, it is expected to further enrich the benchmark to facilitate the pre-train and fine-tune of large model. We hope our proposed large-small model paradigm and perspectives can inspire future work, particularly for downstream tasks with limited resources.

%This paper focuses on addressing the dual challenges currently faced in urban waterlogging detection: the scarcity of annotated data and the poor generalizability of existing methods to recognize the adeptly disguised standing water in real-world settings. We construct a highly diverse urban waterlogging dataset that encompasses a range of adverse conditions not covered by existing datasets, thus facilitating future research development and progression towards practical deployment.

% ---- Acknowledgements ----
\section*{Acknowledgements}
This work was partially supported by National Key R\&D Program of China (2021YFB3100800), National Natural Science Fund of China (62271090, 61771079), Chongqing Natural Science Fund (cstc2021jcyj-jqX0023) and National Youth Talent Project. This work is also supported by Huawei computational power of Chongqing Artificial Intelligence Innovation Center.

\bibliographystyle{splncs04}
\bibliography{main}

% ---- Appendix ----
\clearpage
\section*{A. Additional details on algorithm.}
In \cref{LSM}, we summarize the detailed training process of our proposed LSM-Adapter using a two-stage training strategy.
\begin{algorithm}
   \caption{Training Procedure of LSM-Adapter with two-stage training strategy.}
   \label{LSM}
    {\bfseries input:} Training set $\mathcal{D}_{tr}$.\\
    {\bfseries output:} LSM-Adapter $\mathcal{M}$.
\begin{algorithmic}[1]
   \STATE /* Stage-one */
   \STATE Freeze the image encoder of the large model $\mathcal{M}_{\mathrm{large}}$
   \WHILE {\textit{the maximal iterations are not reached}}
   \STATE $\{x_i,y_i\}_{i=1}^{n}\in{\mathcal{D}_{tr}}$ // \texttt{sample mini-batch}
   \STATE Obtain predictions of small model  $M_{\mathrm{small}}=\mathcal{M}_{\mathrm{small}}(x_i)$
   \STATE Obtain predictions of large model  $M_{\mathrm{large}}=\mathcal{M}_{\mathrm{large}}(x_i)$
   \STATE Update $\mathcal{M}_{\mathrm{small}}$ and $\mathcal{M}_{\mathrm{large}}$ by minimizing the loss function of small model and \cref{eq9}, respectively.
   \ENDWHILE
   \STATE /* Stage-two */
   \STATE Load well-trained $\mathcal{M}_{\mathrm{small}}$ and $\mathcal{M}_{\mathrm{large}}$
   \STATE Freeze $\mathcal{M}_{\mathrm{small}}$, and freeze the image encoder and HE-Adapt of $\mathcal{M}_{\mathrm{large}}$
   \WHILE {\textit{the maximal iterations are not reached}}
   \STATE $\{x_i,y_i\}_{i=1}^{n}\in{\mathcal{D}_{tr}}$ // \texttt{sample mini-batch}
   \STATE $Z=\mathrm{LSM\_Encoder}(x_i)$
   \STATE Obtain $e_{\mathrm{spa}}$ by small model $\mathcal{M}_{\mathrm{small}}$ and spatial prompter // \texttt{spatial prompt}
   \STATE Obtain $e_{\mathrm{sem}}$ by \cref{eq1}, \cref{eq2} and \cref{eq3} // \texttt{semantic prompt}
   \STATE Obtain $e_{\mathrm{sty}}$ by \cref{eq4}, \cref{eq5}, and a convolutional block // \texttt{style prompt}
   \STATE Obtain $e_{\mathrm{p}}$ by \cref{eq6} // \texttt{final prompt}
   \STATE Obtain predictions $M=\mathrm{LSM\_Decoder}(Z,e_{\mathrm{p}})$
   \STATE Update $\mathcal{M}$ by minimizing \cref{eq10}
   \ENDWHILE
\end{algorithmic}
\end{algorithm}

\section*{B. Additional experimental results.}
\subsection*{B.1 Additional implementation details.}
\label{sec:implementation}
For the one-stage training, batch size is set to 2, and the large model and small model are optimized jointly using the AdamW optimizer. The learning rate of the large model is set to 0.0005. For three different small models, i.e., Mask R-CNN~\cite{maskrcnn}, U2Net~\cite{u2net}, and SINet~\cite{sinet}, the learning rate is set to 0.0005, 0.001, and 0.0005, respectively. Cosine annealing decay is applied, and the epoch is set to 40.

For the two-stage training, the large and small models are first optimized individually. The training settings for the large model are the same as in the one-stage training. Mask R-CNN~\cite{maskrcnn} is trained using the SGD optimizer with a learning rate of 0.001, a batch size of 8, and an epoch of 40. U2Net~\cite{u2net} is trained using the Adam optimizer with a learning rate of 0.001, a batch size of 8, and an epoch of 40. And SINet~\cite{sinet} is trained using the Adam optimizer with a learning rate of 0.0001, a batch size of 16, and an epoch of 100. In the second stage, the epoch is set to 20, and the rest of the settings for the large model are the same as in the first stage.

\subsection*{B.2 Model efficiency.}
We evaluate the efficiency of LSM-Adapter$_\mathrm{S}$, including parameter amount and inference time of a single image in \cref{tab:efficiency}. Due to the introduction of the small model, LSM-Adapter$_\mathrm{S}$ has an increased overall parameter amount compared to SAM Adapter~\cite{sam-adapter}. However, the trainable parameters in the second training stage have only increased by 1M, and the inference time per image is comparable to that of the SAM Adapter$_\mathrm{S}$.
%The superscript $\mathrm{s2}$ indicates the second stage of training.

%\begin{table}[t]
\begin{center}
    \begin{minipage}[b]{0.58\linewidth}
    \centering
    \captionof{table}{Model efficiency.}
    \label{tab:efficiency}
    %\setlength{\tabcolsep}{1.2mm}
    %\scalebox{0.6}{
    \resizebox{\textwidth}{!}{
	\begin{tabular}{c|cc}
        \hline
        Method &Params (train / total) &Inference time \\
        \hline
        %SINet    &   &  \\
        SAM-Adapter &4.1M / 93.8M   &0.2088s  \\
        SAM-Adapter$_\mathrm{S}$  &4.1M / 120.8M   &0.3034s  \\
        \hline
        LSM-Adapter$_\mathrm{S}$  &5.1M / 121.8M   &0.3273s \\
        \hline
        \end{tabular}
	}
    \end{minipage}\hfill
    \begin{minipage}[b]{0.4\linewidth}
    \centering
    \captionof{table}{Ablation of $r$.}
    \label{tab:ablation1}
    %\setlength{\tabcolsep}{0.25mm}
    %\scalebox{0.6}{
    \resizebox{\textwidth}{!}{
	\begin{tabular}{c|cccc}
        \hline
        $r$ &Precision &Recall &F1-Score &IoU \\
        \hline
        0.25  &71.59 &55.00 &62.21 &45.15 \\
        0.50   &79.46 &65.29 &71.68 &55.86 \\
        1.00 &\textbf{79.47} &\textbf{70.57} &\textbf{74.76} &\textbf{59.69} \\
        \hline
        \end{tabular}
        }
    \end{minipage}
\end{center}
%\end{table}

\subsection*{B.3 Additional ablation study.}
\subsubsection{Effect of the scale of dataset.}
To investigate the impact of the scale of the annotated dataset, we randomly select training data with and without waterlogging according to the ratio $r$ and then merge them to form a subset for training. The ratio $r$ is set to 1, 0.5, and 0.25, respectively, where a value of 1 indicates training with the original fully annotated data. \cref{tab:ablation1} shows that models trained with more annotated data perform better on downstream tasks, indicating that although foundation model possesses powerful generalization capabilities, additional annotated data is still needed to help the model adapt to downstream tasks.

%\begin{table}[t]
\begin{center}
    \begin{minipage}[b]{0.42\linewidth}
    \centering
    \captionof{table}{Ablation of $\tau$. }
    \label{tab:ablation2}
    \resizebox{\textwidth}{!}{
	\begin{tabular}{c|cccc}
        \hline
        $\tau$ &Precision &Recall &F1-Score &IoU  \\
        \hline
        0.3 &\textbf{81.50} &66.77 &73.40 &57.98  \\
        0.4 &80.55 &68.66 &\textbf{74.13} &\textbf{58.89}\\
        0.5 &78.00 &\textbf{69.85} &73.70 &58.35  \\
        0.6 &78.73 &69.38 &73.76  &58.42  \\
        0.7 &80.18 &67.92 &73.54  &58.15  \\
        \hline
        \end{tabular}
        }
    \end{minipage}\hfill
    \begin{minipage}[b]{0.52\linewidth}
    \centering
    \captionof{table}{Ablation of $\lambda$.}
    \label{tab:ablation3}
    \resizebox{\textwidth}{!}{
	\begin{tabular}{c|cccc}
        \hline
        $\lambda$ &Precision &Recall &F1-Score &IoU \\
        \hline
        10 &52.00 &\textbf{68.15} &58.99 &41.83  \\
        1  &63.88 &62.58 &\textbf{63.22} &\textbf{46.22} \\
        0.1 &63.22 &59.76 &61.44 &44.34  \\
        0.01 &\textbf{70.50} &51.42 &59.47 &42.31  \\
        \hline
        \end{tabular}
	}
    \end{minipage}
\end{center}
%\end{table}

\subsubsection{Effect of Hyper-parameter selection.}
We analyze the impact of two hyper-parameters on performance, the threshold $\tau$ for spatial prompt of point type and the coefficient $\lambda$ in the total loss (\cref{eq8}) of the one-stage training. As the threshold $\tau$ varies from 0.3 to 0.7, the overall performance of the model changes minimally, reaching optimal performance at a threshold of 0.4. The choice of coefficient $\lambda$ significantly affects the model's performance. When SINet~\cite{sinet} is used as the small model and $\lambda$ is set to 1, the overall performance of LSM-Adapter$_\mathrm{S}$ is optimal. For other values, both the F1-score and IoU decline noticeably. Therefore, we recommend careful selection of the coefficient value when choosing other models as task-specific small models to achieve better performance. Generally, the coefficient should be chosen to keep the loss values of the large and small models within the same order of magnitude.

\subsection*{B.4 Experiments on additional datasets.}
We use four additional datasets, CHAMELEON~\cite{chameleon}, CAMO~\cite{camo}, COD10K~\cite{sinet} for camouflaged object detection, and ISTD~\cite{istd} for shadow detection to evaluate the proposed method on more downstream tasks. \cref{tab:other-task} shows that LSM-Adapter$_\mathrm{S}$ outperforms other methods on all datasets.

\begin{table}[tb]
\caption{Experimental results on additional datasets.}
\label{tab:other-task}
%\setlength{\tabcolsep}{1.22mm}
%\scalebox{0.49}{
\centering
\resizebox{\linewidth}{!}{
\begin{tabular}{c|cccc|cccc|cccc|c}
\hline
\multirow{2}{*}{Method} & \multicolumn{4}{c|}{CHAMELEON~\cite{chameleon}}    & \multicolumn{4}{c|}{CAMO~\cite{camo}}         & \multicolumn{4}{c|}{COD10K~\cite{sinet}}       & ISTD~\cite{istd}  \\
\cline{2-14}
& $S_\alpha\uparrow$  & $E_\phi\uparrow$  & $F_\beta^\omega\uparrow$  & $\mathrm{MAE}\downarrow$
& $S_\alpha\uparrow$  & $E_\phi\uparrow$  & $F_\beta^\omega\uparrow$  & $\mathrm{MAE}\downarrow$
& $S_\alpha\uparrow$  & $E_\phi\uparrow$  & $F_\beta^\omega\uparrow$  & $\mathrm{MAE}\downarrow$ &BER$\downarrow$\\
\hline
Mask RCNN~\cite{maskrcnn}
&0.771 &0.798 &0.659 &0.050 &0.680 &0.676 &0.515 &0.107 &0.714 &0.698 &0.521 &0.044 &4.19 \\
U2Net~\cite{u2net}
&0.830 &0.877 &0.699 &0.059 &0.642 &0.690 &0.488 &0.140 &0.688 &0.747 &0.457 &0.076 &3.65 \\
SINet~\cite{sinet}
&0.888 &0.942 &0.816 &0.030 &0.820 &0.882 &0.743 &0.070 &0.815 &0.887 &0.680 &0.037 &2.35 \\
%SAM-Adapter$_\mathrm{b}$
SAM-Adapter~\cite{sam-adapter}
&0.834 &0.858 &0.680 &0.055 &0.800 &0.816 &0.657 &0.094 &0.820 &0.856 &0.657 &0.044 &1.65 \\
%SAM-Adapter$_\mathrm{h}$
%&\underline{0.896} &0.919 &\underline{0.824} &0.033 &\textbf{0.847} &0.873 &\textbf{0.765} &\underline{0.070} &\textbf{0.883} &\textbf{0.918} &\textbf{0.801} &\textbf{0.025} &\textbf{1.43}  \\
\hline	
LSM-Adapter$_\mathrm{M}$
&0.843 &0.893 &0.765 &0.040 &0.784 &0.849 &0.723 &0.081 &0.817 &0.883 &0.710 &0.035 &2.02 \\
LSM-Adapter$_\mathrm{U}$
&0.868 &0.906 &0.740 &0.044 &0.723 &0.762 &0.596 &0.115 &0.778 &0.821 &0.580 &0.053 &2.22  \\
LSM-Adapter$_\mathrm{S}$
%&\textbf{0.903} &\textbf{0.955} &\textbf{0.836} &\textbf{0.024} &\underline{0.825} &\textbf{0.889} &\underline{0.756} &\textbf{0.066} &\underline{0.839} &\underline{0.903} &\underline{0.727} &\underline{0.031} &\underline{1.55}  \\
&\textbf{0.903} &\textbf{0.955} &\textbf{0.836} &\textbf{0.024} &\textbf{0.825} &\textbf{0.889} &\textbf{0.756} &\textbf{0.066} &\textbf{0.839} &\textbf{0.903} &\textbf{0.727} &\textbf{0.031} &\textbf{1.55}  \\
\hline
\end{tabular}
}
\end{table}

\subsection*{B.5 Additional qualitative results.}
\begin{figure}[tb]
  \centering
  \includegraphics[width=\columnwidth]{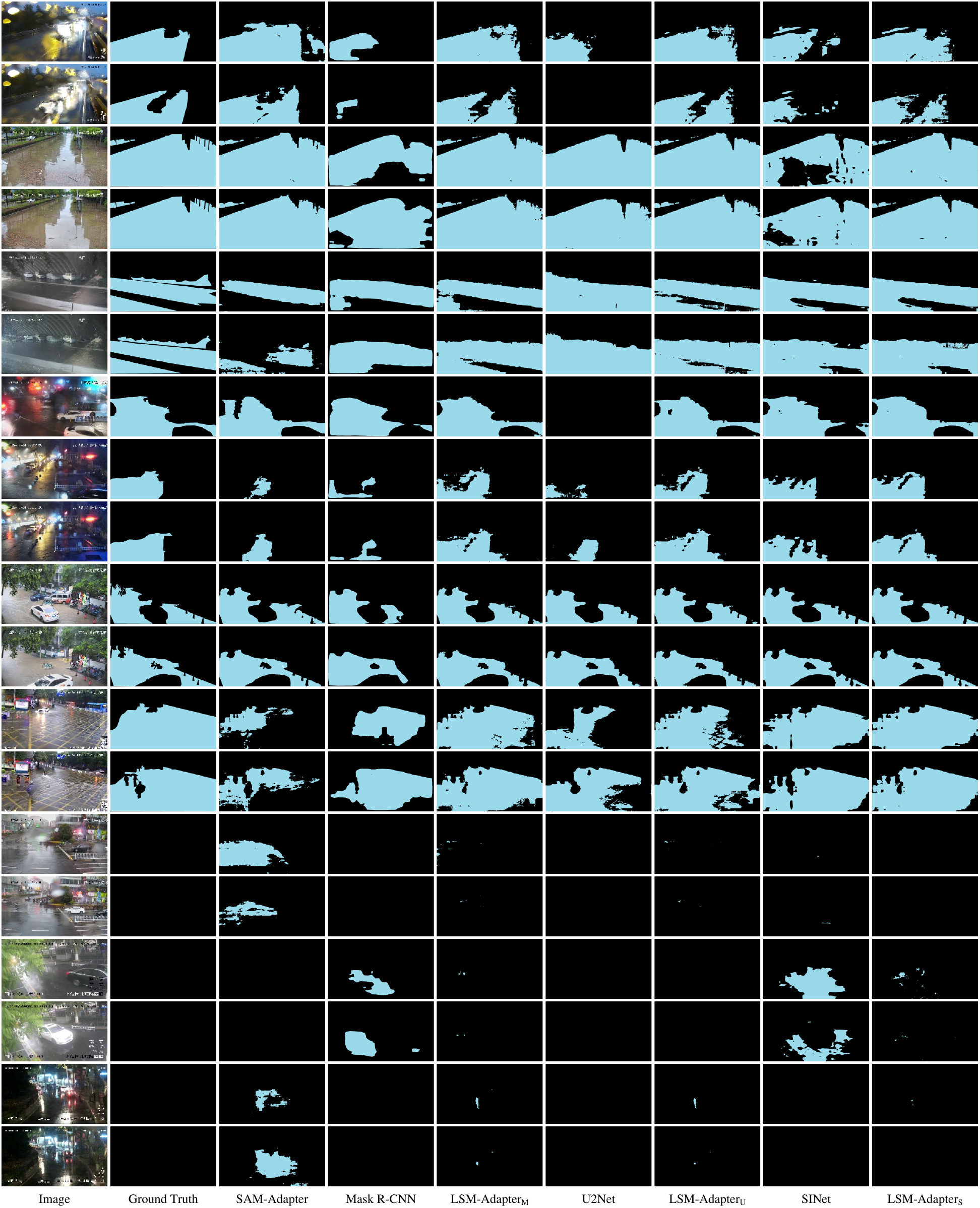}
  \caption{Visual comparison results of different models. M, U, S denotes Mask R-CNN, U2Net and SINet is used as small model, respectively.
  }
  \label{visual2}
\end{figure}

We provide more qualitative results to further demonstrate the effectiveness and superiority of the proposed LSM-Adapter. \cref{visual2} illustrates the visual comparison results between LSM-Adapter and other existing methods, including Mask R-CNN~\cite{maskrcnn}, U2Net~\cite{u2net}, SINet~\cite{sinet}, and SAM-Adapter~\cite{sam-adapter}. LSM-Adapter selects masks from three different small models as spatial prompts and utilizes the two-stage training strategy. It is evident that LSM-Adapter has the prediction results closest to the ground truth, whether compared with the small or large models, even in the case of challenging samples, such as the first, second, eighth, and ninth rows. Meanwhile, we observe that spatial prompts derived from the small model can compensate for the knowledge gap of the large model when the prediction output of the small model is superior, resulting in LSM-Adapter's predictions that more closely align with the ground truth. Concurrently, even when the small model's inferior predictive masks are employed as prompts, LSM-Adapter remains relatively unaffected, maintaining a prediction quality comparable to, if not slightly superior to, that of the standalone large model. %This can be attributed to the semantic prompts, style prompts, and the dynamic prompt combiner's contribution to mitigating the influence of the spatial prompts.

\end{document}